\pdfoutput=1

\documentclass[11pt]{article}

\usepackage[preprint]{acl}

\usepackage{times}
\usepackage{latexsym}

\usepackage[T1]{fontenc}

\usepackage[utf8]{inputenc}

\usepackage{microtype}

\usepackage{inconsolata}

\usepackage{adjustbox}
\usepackage{booktabs}
\usepackage{float}
\usepackage{algorithm}
\usepackage{algorithmic}
\usepackage{amsmath, amssymb}
\usepackage{amsfonts}
\usepackage{mathtools}
\usepackage[normalem]{ulem}
\useunder{\uline}{\ul}{}
\usepackage{multirow}

\title{STEP: Staged Parameter-Efficient Pre-training for Large Language Models}

\author{
    Kazuki Yano${}^{1}$\quad
    {\bf Takumi Ito${}^{1, 2}$\quad
    Jun Suzuki${}^{1,3,4}$}
    \\
    ${}^{1}$Tohoku University\quad
    ${}^{2}$Langsmith Inc.\quad
    ${}^{3}$RIKEN \quad
    ${}^{4}$ NII LLMC \\
    \texttt{yano.kazuki@dc.tohoku.ac.jp}\\
    \texttt{\{t-ito, jun.suzuki\}@tohoku.ac.jp}
    }

\begin{document}
\maketitle
\begin{abstract} 
Pre-training large language models (LLMs) faces significant memory challenges due to the large size of model parameters.
We introduce STaged parameter-Efficient Pre-training (STEP), which integrates parameter-efficient tuning techniques with model growth. 
We conduct experiments on pre-training LLMs of various sizes and demonstrate that STEP achieves up to a 53.9\% reduction in maximum memory requirements compared to vanilla pre-training while maintaining equivalent performance. 
Furthermore, we show that the model by STEP performs comparably to vanilla pre-trained models on downstream tasks after instruction tuning. 
\end{abstract}

\section{Introduction}
\label{sec:introduction}
Large Language Models (LLMs) have become an indispensable foundational technology in artificial intelligence.
Recent LLM development trends, based on scaling laws~\citep{kaplan2020scaling}, involve pre-training Transformer models with a vast number of parameters on massive datasets~\cite{brown2020language}.
Consequently, the pre-training of LLMs requires substantial computational resources, typically involving thousands of GPUs~\citep{touvron2023llama}.
This enormous computational demand presents a significant obstacle to LLM research.

To tackle this challenge, we consider methods for reducing the computational demand in LLM pre-training. 
While there are various approaches to reducing this, we introduce a pre-training method that maintains performance equivalent to vanilla pre-training while constraining the maximum GPU memory requirements to a predetermined threshold.
Specifically, our approach combines model growth~\citep{chen-etal-2022-bert2bert, wang2024lemon} through layer addition with parameter-efficient tuning techniques~\citep{hu2022lora}, which are commonly used in fine-tuning.
\begin{figure}[t]
\centering
\includegraphics[width=1\columnwidth]{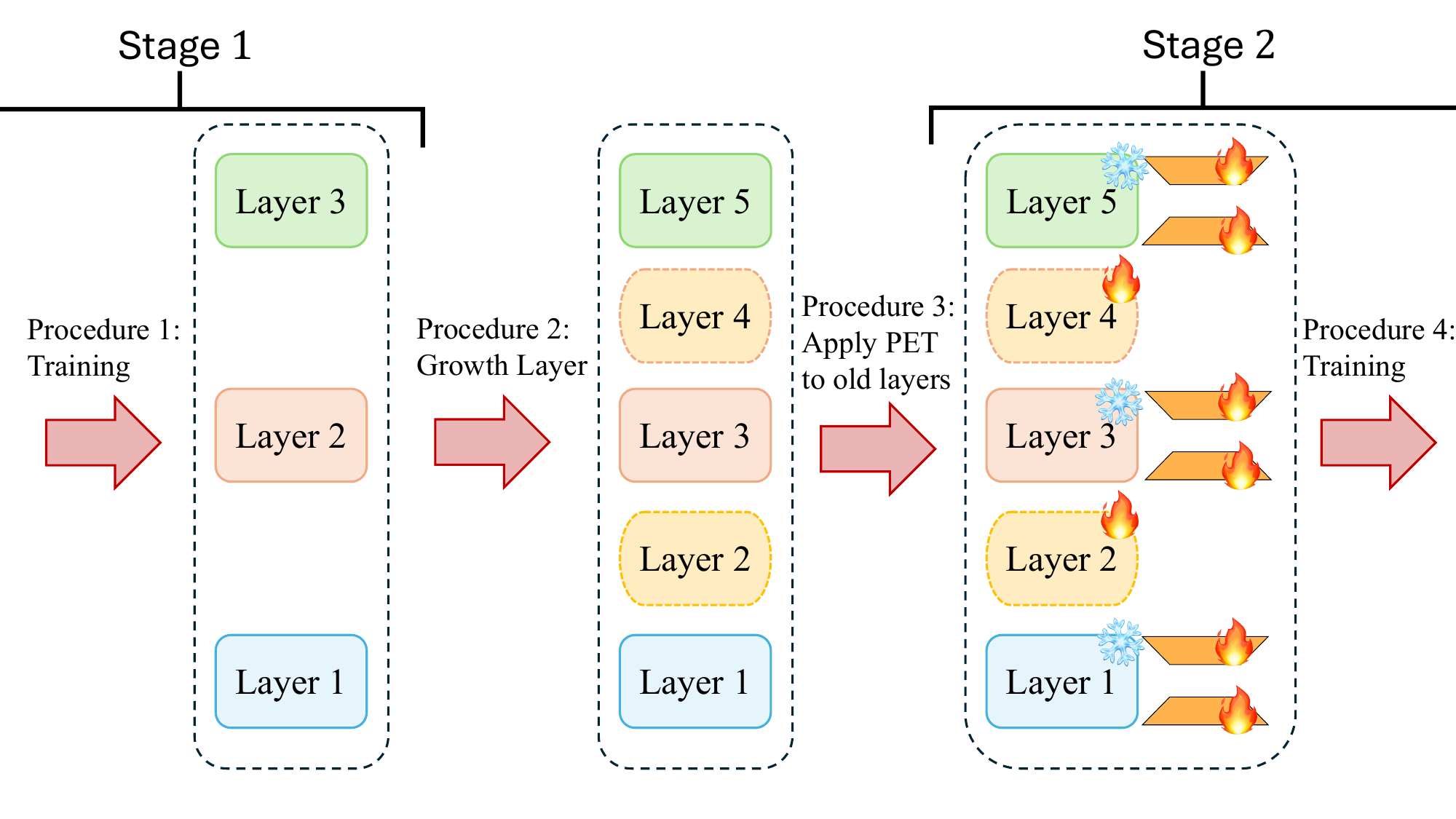}
\caption{Overview of STEP (STaged parameter Efficient Pre-training). First, vanilla pre-training is performed on a small-scale model (Procedure 1). Subsequently, new layers are added to grow the pre-trained model (Procedure 2). The parameters of the pre-trained layers are then frozen, and Parameter-Efficient Training (PET) is applied for alternative training (Procedure 3), followed by retraining of the expanded model (Procedure 4). In Procedure~4, only the parameters added through layer expansion and the small-scale parameters introduced by PET are subject to training.}
\label{fig:step} 
\end{figure}
For a detailed explanation of the proposed method, Figure~\ref{fig:step} presents an overview of our procedure.
Our approach formulates the maximum memory requirements for each stage of the sequential model growth as an integer programming problem, using model configurations as variables. 
We solve this optimization problem to determine the optimal model configurations for each stage, thereby controlling model growth settings to minimize peak memory usage prior to pre-training execution.
This approach enables pre-training while maintaining memory requirements within a predetermined threshold.
Hereafter, we refer to our method as STaged parameter Efficient Pre-training (STEP).
We demonstrate that STEP achieves up to a 53.9\% reduction in maximum memory requirements compared to vanilla pre-training while maintaining equivalent perplexity and performance on domain-specific tasks.
Furthermore, we verify that STEP does not negatively affect the performance of downstream tasks by demonstrating that STEPed models perform on par with the vanilla pre-trained model.

\section{Related Work}
Several memory-efficient training approaches have been actively developed in the literature of training LLMs~\citep{rajbhandari2020zero,korthikanti2023reducing}.
One of the primary approaches involves reducing the number of trainable parameters.
Notable examples include Parameter-Efficient Tuning (PET) methods such as Adapter~\cite{houlsby2019parameter} and LoRA~\cite{hu2022lora}.
Meanwhile, to reduce FLOPs during pre-training, model growth techniques have been proposed~\cite{chen-etal-2022-bert2bert,pan2024preparing}, where training begins with a small-scale model and continues as the model parameters are gradually expanded.
Our proposed method aims to achieve memory-efficient pre-training by appropriately combining PET and model growth techniques.

\paragraph{Paremeter-efficient Tuning.}
PET has primarily been developed for fine-tuning LLMs.
For instance, LoRA is a technique that adds new adapters (low-rank matrices) while keeping the pre-trained LLM parameters frozen, and only trains these adapters.
Since adapters typically contain few parameters, training can be accomplished with minimal memory requirements.

PET is now being applied to pre-training applications.
Here, we describe two representative methods: ReLoRA~\cite{lialin2024relora} and GaLore~\cite{zhao2024galore}.
ReLoRA is a method for pre-training LLMs using LoRA.
A distinctive feature of ReLoRA is that it begins with vanilla pre-training and transitions to LoRA during the training process.
Consequently, from a peak memory requirement perspective, ReLoRA requires the same amount of memory as vanilla pre-training.
GaLore is a method that leverages the low-rank structure of gradients to reduce optimizer states while maintaining performance equivalent to vanilla pre-training.
Unlike ReLoRA, GaLore operates with low memory requirements throughout the entire training process.
These methods can reduce memory usage compared to vanilla pre-training, but they slightly underperform.
\paragraph{Growing pre-trained model.}
Recent studies have shown that growing a smaller model and then continuing to train the larger model can achieve comparable performance with fewer FLOPs compared to training a large model from scratch~\citep{shen2022staged,chen-etal-2022-bert2bert, pan2024preparing}.
In these methods, the operation of increasing the model size is called the Growth Operator, expanding the dimensions of Transformer~\citep{vaswani2017transformer} layers and adding new layers.
Since existing studies train the full parameters of the model, this approach does not reduce the maximum memory requirements.

\section{STEP: STaged parameter Efficient Pre-training}
\subsection{Procedure}\label{subsec:step_procedure}
The following four procedures are an overview of STEP and how it efficiently trains LLMs; 
\paragraph{(Procedure 1)} STEP performs a vanilla pre-training on a model with a much smaller size than the target model size as an initial model.
\paragraph{(Procedure 2)} STEP expands the layers of the initial model to increase its size using the Growth Operator.
\paragraph{(Procedure 3)} STEP also introduces the PET parameters given by the parameter-efficient adaptors for the layers trained in Procedure 1.
\paragraph{(Procedure 4)} STEP continues to pre-train the parameters in layers newly added in Procedure 2 and the adaptors added in Procedure 3 while freezing those in layers trained in Procedure~1. 

After finishing Procedure 4, we obtain the pre-trained model, or we can continue growing the layers by repeating Procedures 2 to 4, alternatively.
Note that the first to fourth red right-arrows in Figure~\ref{fig:step} corresponds to Procedures 1 to 4, respectively.

We select Interpolation used in~\citet{chang2018multilevel,pmlr-v119-dong20c,li2022autoprog} as the Growth Operator in Procedure~2, which adds new layers between existing layers.%
\footnote{We discuss more detailed initialization of the new layers in Appendices~\ref{appendix:intialization} and~\ref{appendix:overfitting}.}
Moreover, we select the low-rank adaptation method~\citep{hu2022lora, lialin2024relora} as PET parameters for performing Procedure~3.

\subsection{Maximum memory requirement of STEP}\label{subsubsec:mem_consis}
We assume that the maximum memory requirement during the pre-training can be estimated by the size of model states, which include model parameters, gradients, and optimizer state.%
\footnote{Other memory usages, such as activations, can be reduced using methods like Activation Recomputation~\citep{korthikanti2023reducing}.}
Moreover, we assume that we use a typical Transformer model~\citep{vaswani2017transformer} and the Adam optimizer~\citep{kingma2014adam} with mixed-precision training~\citep{micikevicius2018mixed}.
Specifically, model parameters and gradients are represented in 16-bit floating-point numbers, while optimizer states are represented in 32-bit floating-point numbers.
When the number of parameters in one layer of the Transformer is $P_{\text{layer}}$ and the number of layers in the model is $n$, the memory usage of the model state, expressed in bytes, is given by
\begin{equation}\label{eq:p_trn}
\hspace{-2mm}
\begin{aligned}
        P_{\text{trn}} &= n  (\underbrace{2P_{\text{layer}}}_{\text{model}} \!+\! \underbrace{2P_{\text{layer}}}_{\text{gradient}} \!+\! \underbrace{12P_{\text{layer}}}_{\text{optimizer}}) \\
                        &= 16 n P_{\text{layer}}, 
    \end{aligned} 
\end{equation}
where the Adam optimizer state consists of three parts: model, gradient momentum, and variance.
Regarding the maximum memory requirement for STEP,
let $n_i$ be the number of layers increased in the $i$-th stage from the $i-1$ stage in STEP.
Let $N_i$ represent the total number of layers in the $i$-th stage model: $N_i = \sum_{k=1}^i n_k$, where $N_0=0$.
Moreover, $E(P_{\text{layer}})$ denotes the number of parameters for a single layer, $P_{\text{layer}}$, added by PET.%
\footnote{Appendix~\ref{appendix:mem_consis} discusses examples of $P_{\text{layer}}$ and $E(P_{\text{layer}})$.}
Then, we estimate the maximum memory requirement for the stage $i$, that is, $P^{\text{\tiny STEP}}_i$, as follows:

\begin{equation}\label{eq:model_state}
\begin{aligned}
P^{\text{\tiny STEP}}_i =&
    16n_iP_{\text{layer}} + 2N_{i-1}P_{\text{layer}}
    \\
    &\quad+ 16N_{i-1} E(P_{\text{layer}})  \end{aligned}
\end{equation}
where the $2N_{i-1}P_{\text{layer}}$ represents the number of frozen model parameters already trained in the 1 to $i-1$ stages, 
the $16n_iP_{\text{layer}}$ indicates the number of newly added model parameters with optimization states added in Procedure~2 and the $16N_{i-1} E(P_{\text{layer}}) $ represents the number of PET parameters added in Procedure~3.
Note that Eq.~\ref{eq:model_state} is identical to Eq.~\ref{eq:p_trn} if $i=1$ since $N_0=0$.

Let $L$ be the number of layers for the model that is finally obtained. 
Then, the solution of the following minimization problem can minimize the maximum memory requirement during the pre-training: 
\begin{equation}\label{eq:mem_equal}
 \underset{\{n_1,\dots, n_K\}}{\text{minimize}} \left\{ \max_{i=1,\dots, K} P^{\text{\tiny STEP}}_i\right\}
 \ \ \textrm{s.t.}\ \ L = N_K 
.
\end{equation}
This minimization problem is essentially an integer linear programming (ILP) problem since $n_i$ for all $i$ are non-negative integers.
Thus, we can straightforwardly obtain the solution set $\{n_i\}^K_{i=1}$ by using a standard ILP solver or manual calculation if $K$ is small, e.g., $K=2$.
Typically, $K$ is small, at most $L-1$, and usually stays below $L/4$, ensuring the problem remains computationally tractable.
As a result, the computational cost is negligible compared to LLM pre-training.%
\footnote{More discussions of the complexity of ILP problems for STEP are in Appendix~\ref{appendix:ILP}.}

\section{Experiments}
We investigate whether STEP can perform equivalent to vanilla pre-training for LLMs at the same FLOPs.%
\footnote{The detailed FLOPs computation is in Appendix~\ref{appendix:FLOPs_comp}.}
We also compare ReLoRA~\citep{lialin2024relora} and GaLore~\citep{zhao2024galore} as parameter-efficient pre-training methods in a fair condition.
Furthermore, to verify whether STEP would not negatively affect the performance of downstream tasks, we will perform instruction tuning on both the STEPed model and the vanilla pre-trained model and compare their performance.

\subsection{Evaluation in pre-training}\label{subsec:pretrain}
\paragraph{Datasets and model.}
We used FineWeb-Edu~\citep{penedo2024the} as the pre-training data.
The model configuration follows LLaMA~\citep{touvron2023llama}.
The detailed configurations are shown in Appendix~\ref{appendix:pretraining_config}.
We selected three different model sizes, namely, 368M, 680M, and 1.2B, to examine whether different model sizes lead to different trends.

\paragraph{Evaluation.}
We calculated the perplexities on two held-out validation sets: one from FineWeb-Edu (10M tokens) and the other from Wiki-Text (0.3M tokens)~\citep{merity2017pointer}.
Furthermore, we evaluated the accuracy of several typical downstream tasks for evaluating LLMs.
\footnote{Detailed evaluation settings and tasks are in Appendix~\ref{appendix:evaluation}}

\begin{table}[t]
    \centering
    \small
    \begin{tabular}{ccc}
        \toprule
        Model Size & Hidden &  Layers \\ \midrule
        215M \textrightarrow 368M & 1600  & 7 \textrightarrow 12  \\ 
        396M \textrightarrow 680M & 1536  & 14 \textrightarrow 24  \\ 
        704M \textrightarrow 1.2B & 2048  & 14 \textrightarrow 24  \\ 
        553M \textrightarrow 956M \textrightarrow 1.2B & 2048 & 11 \textrightarrow 19 \textrightarrow 24 \\
        \bottomrule
    \end{tabular}
    \caption{The STEP configurations used in the experiments. The number of parameters and layers for each model at different stages are shown.
    The last row shows a three-stage growth process.}
    \label{tab:model_config}
\end{table}
\begin{figure}[t]
\centering
\includegraphics[width=1\columnwidth]{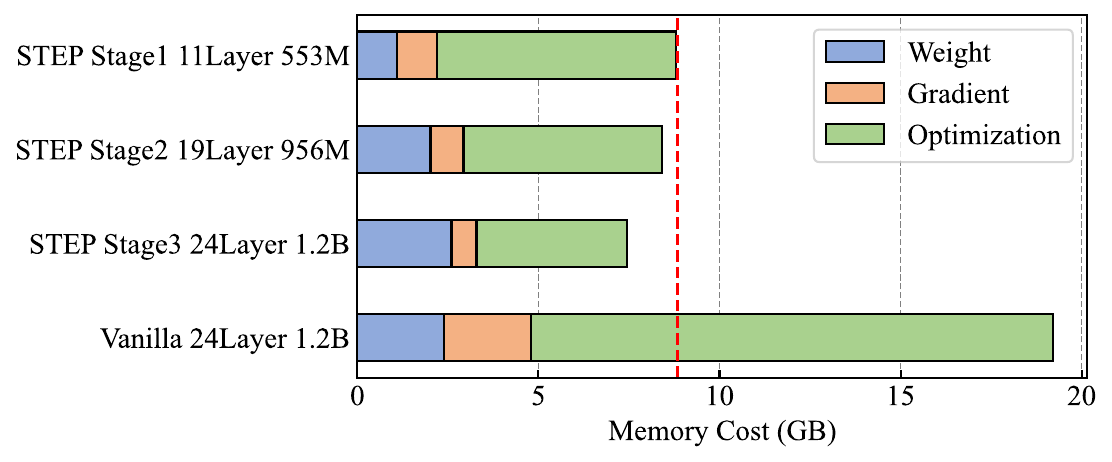}
\vskip -0.11in
\caption{Memory consumption of pre-training 1.2B in Table~\ref{tab:model_config}. 
STEP allows for increasing the model size while keeping memory usage consistent at every stage.
}
\label{fig:memory_breakdown} 
\end{figure}
\begin{table*}[!htb]
    \tabcolsep 1mm
    \centering
    \small
    \begin{tabular}{lccccccccc}
        \toprule
        \multirow{2}{*}{} & \multicolumn{2}{c}{Perplexity \textdownarrow}  & \multicolumn{7}{c}{Accuracy \textuparrow}         \\ \cmidrule(r){2-3} \cmidrule(r){4-10} 
            & Validation & Wikitext  & LAMBADA  & ARC-e & ARC-c  & Winogrande  & PIQA  & OBQA  & HellaSwag  \\ \midrule
            \multicolumn{10}{l}{\hspace{-0.22cm} {\ul 368M}} \\ 
            Vanilla (5.9G) & 16.9 & 32.1 & 29.2 & 52.2 & 27.3 & 50.3 & 64.9 & \textbf{32.4} & 37.3  \\
            ReLoRA (5.9G)  & 17.4 & 33.1 & 28.8 & 51.9 & 27.8 & 50.5 & 65.1 & 31.2 & 36.5      \\
            GaLore (3.3G)  & 21.6 & 43.1 &  22.8 & 48.1 & 25.7 &  \textbf{51.2}  & 62.5   &  30.8  &  31.7   \\
            STEP-2stages (3.4G)    &  \textbf{16.7}  &  \textbf{31.5}  &  \textbf{31.5}  &  \textbf{52.3}  &  \textbf{28.4}  &  49.7 &  \textbf{65.5}   &  32.0   &  \textbf{37.8} \\
            \midrule
            \multicolumn{10}{l}{\hspace{-0.22cm} {\ul 680M}} \\ 
            Vanilla (10.9G) & \textbf{14.6} & \textbf{26.0} & 34.8 & 55.8 & \textbf{30.2} & 52.3 & \textbf{69.7} & \textbf{36.2} & 43.2  \\
            ReLoRA (10.9G)  &  15.1  &  27.3  &  34.0  &  54.1  &  29.0  &  52.1  &  67.3   & 33.8 &  42.1  \\
            GaLore (6.0G)  &  19.4  & 37.5  & 25.0 & 49.1 & 26.2  & 51.4 & 62.4  & 29.6  &  33.8  \\
            STEP-2stages (6.3G)    &  \textbf{14.6}  &  \textbf{26.0}  &  \textbf{35.4} &  \textbf{56.0}  &  29.7   &  \textbf{55.3}  &  67.7  & 34.2   & \textbf{43.7} \\
            \midrule
            \multicolumn{10}{l}{\hspace{-0.22cm} {\ul 1.2B}} \\ 
            Vanilla (19.3G) & \textbf{12.9} & \textbf{22.1} & \textbf{39.9} & 62.0 & 31.1 & 52.1 & 71.0 & 34.6 & 48.8 \\
            ReLoRA (19.3G)  &  13.5  & 23.6 &  37.0  &  60.3  &  31.1  &  51.9  &  70.1  &  34.6  &  46.6   \\
            GaLore (10.4G)  &  17.4 &  35.3  &  28.0  &  51.9  &  26.6  &  50.4  &  65.7  &  32.2  &  36.6  \\
            STEP-2stages (10.6G)    &   \textbf{12.9}  &  22.3   &  39.7   &   \textbf{62.4}  &  \textbf{34.3}    &   \textbf{54.8}   &  70.0  &  35.4   & 48.4 \\
            STEP-3stages (8.9G)    & \textbf{12.9} &  \textbf{22.1}  &  38.7  &  61.0  & 32.7   &    53.8 &  \textbf{71.2}  &  \textbf{35.6}   & \textbf{48.9} \\  
            \bottomrule
\end{tabular}
\caption{Perplexity and accuracy of vanilla pre-training
(Vanilla), ReLoRA, GaLore, and STEP. The numbers in parentheses indicate the maximum memory requirements for each method during pre-training in this experiment.
}
\label{tab:main_result}
\end{table*}

\begin{table*}[t]
    \centering
    \small
    \setlength{\tabcolsep}{3pt} 
    \begin{tabular}{lcccccccc|c}
    \toprule
        & Writing & Roleplay & Reasoning & Math & Coding & Extraction & STEM & Humanities &  Average \\
        \midrule
        Vanilla 1.2B &  2.85 & 3.25 & \textbf{2.60} & 1.10 & 1.00 & 1.10 & 3.20 & 2.75 & 2.26 \\
        \midrule
         STEP-2stages 1.2B & \textbf{3.10} & \textbf{3.95} & 1.95 & 1.00 & 1.05 & 1.10 & \textbf{3.73} & 2.60 & \textbf{2.30} \\
         STEP-3stages 1.2B & 2.85 & 3.30 & 1.95 & \textbf{1.35} & \textbf{1.10} & 1.10 & 3.25 & \textbf{3.20} & 2.26 \\
                              
    \bottomrule
    \end{tabular}
    \caption{Category-specific and average scores on MT-Bench to the answers generated by models
instruction-tuned with vanilla pre-trained models (Vanilla) and STEPed models (STEP-2stages and STEP-3stages).}
\label{tab:instruction_result}
\end{table*}

\paragraph{Configuration of STEP.}
We focus on evaluating STEP when the Growth Layer Operator is applied once during its pre-training, that is, STEP-2stages ($K=2$).
Additionally, we evaluate the STEP-3stages ($K=3$) only for the 1.2B model.

Given the number of layers $L$ with the fixed dimension of hidden layers, we compute $\{n_1, n_2\}$ for STEP-2stages, or $\{n_1, n_2, n_3\}$ for STEP-3stages, respectively, that can minimize the maximum memory requirements by Eq.~\ref{eq:mem_equal}.
Table~\ref{tab:model_config} shows the calculated numbers of layers when the target model sizes are one of \{368\text{M}, 680\text{M}, 1.2\text{B}\}.
Figure~\ref{fig:memory_breakdown} shows an example of memory requirements when the target model size is 1.2B for vanilla pre-training and each stage of the STEP-3stages.

The schedule for applying the Growth Layer Operator is set to occur when $75$\% of the total training steps for each stage have been completed.

\paragraph{Results.}
Table~\ref{tab:main_result} shows the performance of vanilla pre-training, ReLoRA, GaLore, and STEP. 
STEP outperformed both ReLoRA and GaLore.
Additionally, STEP achieved equivalent performance to the vanilla pre-training while significantly reducing the maximum memory requirement from 5.9G to 3.4G (42.3\% reduction), 10.9G to 6.3G (42.2\% reduction), and 19.3G to 8.9G (53.9\% reduction) for 368M, 680M, and 1.2B models, respectively.
Furthermore, the results of STEP-2stages and STEP-3stages at 1.2B parameters show that increasing the number of stages leads to further reduction in memory usage without compromising performance.
These results suggest that STEP can efficiently pre-train LLMs with reduced memory usage.%
\footnote{Appendix~\ref{appendix:analysis} discusses the mechanism behind STEP.}

\subsection{Evaluation in instruction tuning}\label{sebsec:instruction}
\paragraph{Data and evaluation measure.}
For instruction tuning, we used the Alpaca dataset~\citep{alpaca}.
Details of the training configurations are presented in Appendix~\ref{appendix:instruction_config}.
We compare three 1.2B models one trained with vanilla pre-training, while the other two were trained using STEP (STEP-2stages, STEP-3stages).
We evaluate these instruction-tuned models on MT-Bench~\citep{zheng2024judging} by generating model responses to 80 multi-turn questions and assign a numerical rating out of 10 to each response by GPT-4~\citep{achiam2023gpt}.

\paragraph{Results.}
Table~\ref{tab:instruction_result} shows the MT-bench scores of the vanilla pre-trained models (Vanilla) and STEPed models (STEP-2stages and STEP-3stages).
We found that the scores of STEPed models were either equal to or slightly higher than those of the vanilla pre-trained model.
These results indicate that STEP does not have a negative impact on downstream tasks.

\section{Ablation Study}\label{sec:ablation}
We examine the effective position for new layers and the effectiveness of PET, both key components of STEP.
\footnote{The ablation study on the initialization methods for new layers and the schedule of applying the Growth Operator is conducted in Appendix~\ref{appendix:ablation}.}
We used the model settings with a target size of 680M from Table~\ref{tab:model_config}.

\paragraph{Effective position for adding new layers.}
\begin{figure}[t]
\centering
\includegraphics[width=1\columnwidth]{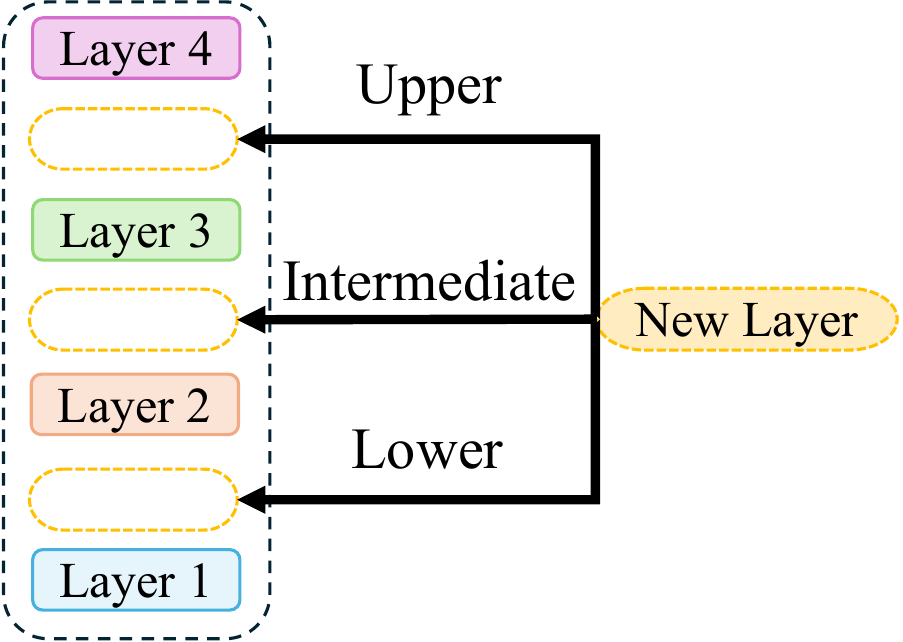}
\caption{Illustration of different strategies for adding new layers in STEP. `Upper' adds layers at the top, `Intermediate' inserts layers in the middle, and `Lower' adds layers at the bottom.}
\label{fig:effective_position} 
\end{figure}
\begin{table}[t]
    \centering
    \small
    \begin{tabular}{lcc}
    \toprule
        & position & \textbf{680M} \\
        \midrule
        Vanilla & & 14.56   \\
        \midrule
        STEP-2stages & Upper        & \textbf{14.56} \\
                    & Intermediate & 14.80 \\
                    & Lower        & 15.06  \\
                    & Random       & 14.82 \\
    \bottomrule
    \end{tabular}
    \caption{Validation perplexities for vanilla pre-trained models (Vanilla) and STEPed model (STEP-2stages) when changing the location of newly added layers.}
    \label{tab:effective_position_result}
\end{table}
We investigated the most effective position for performance improvement when using Interpolation-Mean in Procedure~2 of STEP.
As shown in Figure~\ref{fig:effective_position}, we conducted experiments for Upper, where new layers are added collectively at the top; Intermediate, where they are inserted in the middle; and Lower, where they are added at the bottom.
Additionally, we conducted experiments for Random, where the position of additional layers is determined randomly.

As shown in Table~\ref{tab:effective_position_result}, we can see a trend that performance improves more when layers are added towards the upper part, and this is better than randomly deciding the location for layer addition.

\paragraph{The effect of PET parameters.}
\begin{table}[t]
    \centering
    \small
    \begin{tabular}{lcc}
    \toprule
        &  & \textbf{680M} \\
        \midrule
        Vanilla & & 14.56 (10.9G)  \\
        \midrule
        STEP-2stages & w/ PET        & \textbf{14.56} (6.34G) \\
                    & w/o PET & 14.66 (5.32G)  \\
    \bottomrule
    \end{tabular}
    \caption{Validation perplexities for vanilla pre-trained models (Vanilla) and STEPed model (STEP-2stages) w/ and w/o PET.}
    \label{tab:without_lora_result}
\end{table}
This experiment verifies whether the PET introduced in STEP contributes to performance improvement.
Specifically, we conducted an experiment skipping Procedure 3 in Section~\ref{subsec:step_procedure}.

As shown in Table~\ref{tab:without_lora_result}, PET contributes to performance improvement, and without it, the performance is inferior to the vanilla pre-trained model.

\section{Conclusion}
Pre-training LLM requires substantial memory, posing a challenge for LLM research. 
We proposed a novel training method called STEP, which enables LLM pre-training with reduced memory requirements. 
Our experiments demonstrated the effectiveness of STEP; specifically, STEP achieved equivalent performance to vanilla pre-training and downstream tasks after instruction tuning, while reducing peak memory usage by up to 53.9\%.
We hope our results encourage researchers who aim to engage in LLM pre-training research but have only limited computing resources.

\clearpage
\section*{Limitations}
Several limitations of our study should be addressed in future research.
First, our experiments have been limited to the FineWeb-Edu dataset and only LLaMA architecture.
We need to see if the results can be replicated on other pre-training datasets and other architectures.
Second, our experiments focused on relatively smaller model sizes compared to the recent LLMs with billions of parameters, such as those with 7B or more. 
Third, since STEP begins training with smaller models, it requires a larger amount of training tokens at the same FLOPs of vanilla pre-training. 
While we conducted experiments in situations where the training corpus is unconstrained, the effectiveness of STEP in data-constrained situations remains unexplored.
Finally, this paper focuses its experiments on Transformers, as they are the most commonly used architecture for LLMs. However, the potential applicability to other architectures, such as State Space Models~\citep{gu2024mamba}, has not been verified in this study. 

\section*{Ethical Considerations}
We exclusively used publicly available datasets for pre-training, fine-tuning, and evaluation. 
Moreover, we developed the language models entirely from scratch, avoiding the use of any publicly available models. 
Given that our proposal is a framework for pre-training language models, the risk of ethical concerns is minimal.

\section*{Acknowledgements}

This work was supported by the ``R\&D Hub Aimed at Ensuring Transparency and Reliability of Generative AI Models'' project of the Ministry of Education, Culture, Sports, Science and Technology, and JST Moonshot R\&D Grant Number JPMJMS2011-35 (fundamental research).

In this research work, we used the ``mdx: a platform for building data-empowered society''.


\clearpage
\appendix

\section{The initialization of the new layer}\label{appendix:intialization}
When using Interpolation, most existing studies~\citep{shen2022staged, li2022autoprog, wu2024llama} have adopted the method of copying weights from lower layers to initialize new layers, specifically $\phi^{\mathrm{new}}_{2i} = \phi^{\mathrm{new}}_{2i-1} = \phi_i$, which we call Interpolation-Copy. 
On the other hand, bert2BERT~\citep{chen-etal-2022-bert2bert} proposed a method to expand the width by not only copying from lower layers but also mixing weights copied from both lower and upper layers, demonstrating improved performance compared to simple copying from lower layers.
Inspired by this, we further extend Interpolation by incorporating an idea of a fusing method that averages the parameters of the two layers \citep{pmlr-v157-o-neill21a}, namely,
$\phi^{\mathrm{new}}_{2i} = (\phi_i + \phi_{i+1})/{2}$, which we call Interpolation-Mean.
\citet{shen2022staged, wu2024llama} apply zero-initialization, called function preserving initialization (FPI), to some modules when applying Interpolation to preserve the loss value.
However, as \citet{yao2024masked} points out, the existing layers may receive gradients similar to the previous stage, leading to unnecessary constraints and potentially slowing down the convergence of the model.
Therefore, we do not use FPI.
The validity of these settings will be
verified through experiments.

\section{Overfitting in smaller initial models}\label{appendix:overfitting}
Although there might be concerns about overfitting in the STEP method due to initial training on smaller models, according to Kaplan's Scaling Law~\citep{kaplan2020scaling}, overfitting can be mitigated with sufficient data.
Given that pre-training of large language models typically involves vast amounts of data, this abundance of data in LLM pre-training scenarios theoretically minimizes overfitting risks.

\section{STEP with LLaMA and LoRA}\label{appendix:mem_consis}

In STEP, we use ReLoRA for PET and LLaMA as the model. When not considering Grouped Query Attention~\citep{ainslie2023gqa} in LLaMA, the Self-Attention layer contains four matrices of size ($d_{\text{hidden}}$, $d_{\text{hidden}}$). 
Additionally, the FFN layer has three matrices of size ($\frac{8}{3}d_{\text{hidden}}$, $d_{\text{hidden}}$), and there are two vectors of size $d_{\text{hidden}}$ for Layer Normalization. Therefore, $P_{\text{layer}}$ is given by:
\begin{equation}
    \begin{aligned}
         P_{\text{layer}} &= 4d^2_{\text{hidden}} + 3 \times \frac{8}{3}d^2_{\text{hidden}} + 2d_{hidden} \\
                        &= 12d^2_{\text{hidden}} + 2d_{\text{hidden}}
    \end{aligned}
\end{equation}
Furthermore, since ReLoRA assigns two matrices of size ($d$, $r$) to a matrix of size ($d$, $d$), we have:
\begin{equation}
    \begin{aligned}
    E(P_{\text{layer}}) &= 8 (rd_{\text{hidden}}) + 3r(d_{\text{hidden}} + \frac{8}{3}d_{\text{hidden}}) \\
                     &= 19rd_{\text{hidden}}
    \end{aligned}
\end{equation}

\section{Complexity of ILP Problems}\label{appendix:ILP}

The integer linear programming (ILP) used in STEP is not particularly complex. 
The upper bound on the number of growth stages is the final number of layers, $L$, e.g., $L=24$. 
In practical applications, the number of growth stages, $K$, is typically small (e.g., $K=2$ or $K=3$, or at most around $L/4$). 
This results in a relatively small number of variables, which helps limit the problem's complexity. 
In our experiments using an integer programming solver, we obtained solutions within 2 or 3 seconds for cases where $K\approx 10$, though actual speed may vary depending on the performance of the hardware and the solver's implementation. 
Therefore, the computational cost is negligible compared to the LLM pre-training, which takes at least several hours and is not a significant concern.

\section{FLOPs Computation}\label{appendix:FLOPs_comp}
Let $C$ be the FLOPs, $N$ the number of non-embedding parameters, and $T$ the total number of tokens used in training. 
Then, $C \approx 6NT$.
The coefficient 6 represents the number of floating point operations required for one step, consisting of 2 floating point operations for the forward pass and 4 floating point operations for other calculations such as the backward pass.
Therefore, if we denote the number of trainable parameters as $N_{\text{trainable}}$ and the number of frozen, untrainable parameters as $N_{\text{untrainable}}$, the FLOPs can be calculated as $C \approx (6N_{\text{trainable}} + 2N_{\text{untrainable}})T$.

\section{Details of pre-training configurations}\label{appendix:pretraining_config}
\begin{table}[ht]
\centering
\small
\resizebox{\linewidth}{!}{%
\begin{tabular}{lc}
\toprule
\textbf{Configurations}     & \textbf{Selected Value} \\ \midrule
\multicolumn{2}{l}{\hspace{-0.22cm} {\ul \textit{Common settings}}}  \\
Optimizer                   & AdamW ($\beta_1 = 0.9, \beta_2 = 0.95$) \\
Weight decay & 0.1 \\
Learning rate schedule      &  cosine  \\
Warmup steps                & 1000 \\ 
Seq. len. & 1024 \\ \midrule
\multicolumn{2}{l}{\hspace{-0.22cm} {\ul \textit{ReLoRA settings}}} \\ 
LoRA rank & 128 \\
ReLoRA reset & 5000 \\
Restart warmup steps        & 500 \\ \midrule
\multicolumn{2}{l}{\hspace{-0.22cm} {\ul \textit{GaLore settings}}}\\ 
GaLore rank & 128 \\
Update projection gap & 200 \\
Galore scale & 0.25 \\
\bottomrule
\end{tabular}%
}
\caption{List of training configurations common to all model sizes in pre-training experiments in Section~\ref{subsec:pretrain}.}
\label{tab:training_config_all}
\end{table}
\begin{table*}[!htb]
    \centering
    \small
    \resizebox{\linewidth}{!}{%
    \begin{tabular}{lcccccc}
        \toprule
            & Learning rate  & Learning rate schedule & Batch size  & Training tokens & Training steps & FLOPS  \\ \midrule
            \multicolumn{6}{l}{\hspace{-0.22cm} {\ul 368M}} \\ 
            Vanilla  & 5e-4 & cosine & 360K & 7B & 20K & 1.63e+19 \\
            ReLoRA  & 5e-4 & cosine restarts & 360K & 13B & 40K &  1.63e+19      \\
            GaLore  & 1e-2 & cosine & 360K & 7B & 20K &  1.63e+19 \\
            STEP-2stages &  5e-4  &  cosine &  360K  & 11B & 33K &  1.63e+19 \\
            \midrule
            \multicolumn{6}{l}{\hspace{-0.22cm} {\ul 680M}} \\ 
            Vanilla & 4e-4 & cosine & 688K & 14B & 20K &  5.55e+19  \\
            ReLoRA &  4e-4 & cosine restarts & 688K & 23B & 43K &  5.55e+19 \\
            GaLore  &  1e-2  & cosine & 688K & 14B & 20K &  5.55e+19  \\
            STEP-2stages &  4e-4  & cosine &  688K &  21B  & 33K &   5.55e+19\\
            \midrule
            \multicolumn{6}{l}{\hspace{-0.22cm} {\ul 1.2B}} \\ 
            Vanilla  & 3e-4 & cosine & 1179K & 24B & 20K &  1.73e+20 \\
            ReLoRA   &  3e-4 & cosine restarts &  1179K  &  43B  & 43K &  1.73e+20  \\
            GaLore &  1e-2 &  cosine  &  1179K  &  24B  & 20K &   1.73e+20  \\
            STEP-2stages  & 3e-4 &  cosine &  1179K  &  39B  & 33K &  1.73e+20 \\
            STEP-3stages & 3e-4 & cosine  &  1179K & 53B & 43K &  1.73e+20 \\  
            \bottomrule
\end{tabular}%
}
\caption{Hyperparameters specific to each model
setting and method in Table~\ref{tab:main_result}. Batch size is specified in tokens.
}
\label{tab:training_config_each}
\end{table*}

We used GPT-2 vocabulary~\citep{radford2019language}, although the architecture is based on LLaMA.
The training configurations common to all model settings (368M, 680M, 1.2B) are shown in Table~\ref{tab:training_config_all}.
The training configurations specific to each model setting are presented in Table~\ref{tab:training_config_each}.
We adhered to the hyperparameter settings reported in the papers for ReLoRA~\citep{lialin2024relora} and GaLore~\citep{zhao2024galore}.

All experiments run on NVIDIA A100 GPUs.
\paragraph{Re-initialization of learning rate scheduler.}
When adding layers in Procedure~2, we reset the optimizer state for old layers by applying PET to those.
Moreover, in Procedure~4, to facilitate more efficient training of the new layers, the learning rate is rewarmed to the value used in Procedure~1.
\section{Evaluation of pre-trained models}\label{appendix:evaluation}
Using the lm-evaluation-harness framework, we
report the acc-norm score to follow ~\citet{brown2020language}.
For language modeling tasks, we evaluated perplexity on the Wiki-text dataset~\citep{merity2017pointer} and accuracy on the LAMBADA dataset~\citep{paperno-etal-2016-lambada}.
We assessed zero-shot performance on various commonsense reasoning tasks, including WinoGrande~\citep{sakaguchi2021winogrande}, PIQA~\citep{piqa}, and HellaSwag~\citep{zellers-etal-2019-hellaswag}.
Additionally, we measured zero-shot performance on question-answering tasks, specifically ARC~\citep{clark2018think} and OBQA~\citep{mihaylov-etal-2018-suit}.
We utilized the lm-evaluation-harness framework~\citep{eval-harness} and reported the acc-norm score to follow ~\citet{brown2020language}.

\section{Details of instruction-tuning configurations}\label{appendix:instruction_config}
\begin{table}[ht]
\centering
\small
\tabcolsep 3pt
\begin{tabular}{@{}lc@{}}
\toprule
\textbf{Configurations}     & \textbf{Selected Value} \\ \midrule
Optimizer                   & AdamW ($\beta_1 = 0.9, \beta_2 = 0.95$) \\
Learning Rate               & 0.0001\\
Learning Rate Schedule      &  cosine \\
Warmup steps        & 100 \\
epoch & 2 \\
\bottomrule
\end{tabular}
\caption{Training configurations in our instruction tuning in Section~\ref{sebsec:instruction}.}
\label{tab:instruction_config}
\end{table}
We show the training configurations used in the instruction tuning in Table~\ref{tab:instruction_config}.
All three instruction-tuned models in Table~\ref{sebsec:instruction} undergo full-parameter tuning.

\section{Extensive ablation study}\label{appendix:ablation}
\paragraph{Initialization of the new layer.}
\begin{table}[t]
    \centering
    \small
    \begin{tabular}{lcc}
    \toprule
        & Interpolation & \textbf{680M} \\
        \midrule
        Vanilla & & 14.56   \\
        \midrule
        STEP-2stages & Copy w/ FPI        & 14.59 \\
                    & Copy w/o FPI & 14.60 \\
                    & Mean w/ FPI        & 14.63  \\
                    & Mean w/o FPI       & \textbf{14.56} \\
    \bottomrule
    \end{tabular}
    \caption{Validation perplexities for vanilla pre-trained models (Vanilla) and STEPed model (STEP-2stages) using different initialization of the new layer.}
    \label{tab:initialization_result}
\end{table}
As described in Section~\ref{appendix:intialization}, we investigate the impact of initialization.
We conducted four experiments, with and without FPI, for both Interpolation-Copy and Interpolation-Mean.
The results of this ablation study are shown in Table~\ref{tab:initialization_result}.
As an overall trend, we can see that using FPI does not lead to significant performance improvements.
We expected Interpolation-Mean to contribute more to performance improvement than Copy, and while this is true when FPI is not used, Interpolation-Mean with FPI showed the most significant performance degradation.
FPI had little impact on performance and actually tended to degrade it, while Interpolation-Mean without FPI demonstrated the best performance results.

\paragraph{The schedule for applying the Growth Layer Operator.}
\begin{table}[t]
    \centering
    \small
    \begin{tabular}{lcc}
    \toprule
        & schedule timing & \textbf{680M} \\
        \midrule
        Vanilla & & 14.56   \\
        \midrule
        STEP-2stages & 100\%        & 14.75 \\
                    & 75\% & \textbf{14.56} \\
                    & 50\%        & \textbf{14.56}  \\
                    & 25\%     & 14.94 \\
    \bottomrule
    \end{tabular}
    \caption{Validation perplexities for vanilla pre-trained models (Vanilla) and STEPed model (STEP-2stages) at different schedule timings.}
    \label{tab:schedule_result}
\end{table}
While in our experiments (Section~\ref{subsec:pretrain}), the Growth Layer Operator was applied at 75\% of the training steps in each stage, this experiment examined the schedule timing in more detail.
Specifically, we conducted four experiments, applying the Growth Layer Operator at 25\%, 50\%, 75\%, and 100\% completion of the training steps.
The experimental results are shown in Table~\ref{tab:schedule_result}. 
As the results indicate, the best performance was achieved at 50\% and 75\% points, while applying the Growth Layer Operator at 25\% and 100\% points showed relatively poor results.
One possible reason for this is that at the 25\% point, the training of each layer has not yet progressed sufficiently, and applying PET to existing layers in this state may dramatically slow down the training of each layer.
Additionally, applying the Growth Layer Operator at the 100\% point may cause the model to escape from local optima due to learning rate rewarm and optimizer state resets, resulting in increased loss and requiring more training steps to converge to a better optimal solution.

\section{Discussion on the mechanisms behind STEP}\label{appendix:analysis}
In this section, we discuss why STEP works sufficiently well. We will focus our discussion on model growth and parameter-efficient tuning, which constitute STEP.
\paragraph{Optimization dynamics of model growth.}
Recent research by~\citet{agarwal2024stacking} has demonstrated that adding layers to the upper part of Transformer layers (a process known as ``stacking'') is particularly effective from an optimization perspective. 
Specifically, this work shows that stacking behaves more like accelerated gradient descent rather than simple gradient descent, enabling more efficient learning. 
This finding could potentially provide theoretical support for STEP's strategy of adding layers primarily to the upper portions of the model.%
\footnote{See Appendix~\ref{appendix:ablation} for this strategy.}
Furthermore, empirical observations reported in~\citet{chen-etal-2022-bert2bert} indicate that attention patterns learned by BERT models trained from scratch are commonly seen across layers. 
This insight helps explain why STEP can effectively learn basic attention patterns in its initial stages with a smaller model and then successfully transfer this knowledge to larger models as they grow.
\paragraph{Local low-rank structure and parameter-efficient tuning.}
The effectiveness of Parameter-Efficient Tuning (PET) methods like LoRA~\citep{hu2022lora} and ReLoRA~\citep{lialin2024relora}, which STEP utilizes, is grounded in the theory of local low-rank structure in neural networks. 
This theory posits that the updates to the weights of a neural network during training often lie in a low-dimensional subspace.
By leveraging this property, PET methods can achieve comparable performance to full fine-tuning while updating only a small number of parameters.
In the context of STEP, this background explains how we can maintain high performance while significantly reducing memory requirements. By applying PET to the layers trained in earlier stages, STEP can continue to update these layers efficiently without the need to store full-rank gradients and optimizer states.

Through these discussions, we can better understand why STEP is able to achieve comparable performance to traditional pre-training methods while significantly reducing memory requirements.

\end{document}